\documentclass{ifacconf}

\usepackage{graphicx}      
\usepackage{natbib}        

\usepackage[utf8]{inputenc}
\usepackage[T1]{fontenc}
\usepackage[nolist]{acronym}
\usepackage{algorithm}
\usepackage{algorithmicx}
\usepackage{algpseudocode}
\usepackage[title]{appendix}
\usepackage{amsmath,amssymb,amsfonts}
\usepackage{bm}
\usepackage{caption}
\usepackage{color}
\usepackage{booktabs}
\usepackage{enumitem}
\usepackage{listings}
\usepackage{tikz}
\usepackage{manyfoot}
\usepackage{multirow}
\usepackage{mathrsfs}
\usepackage{nameref}
\usepackage{siunitx}
\usepackage{url}
\usepackage{subcaption}
\usepackage{textcomp}
\usepackage{xcolor}

\graphicspath{{./figures}}
\newacro{hri}[HRI]{Human-Robot Interaction}
\newacro{hrc}[HRC]{Human-Robot Collaboration}
\newacro{gui}[GUI]{Graphical User Interface}
\newacro{pa}[PA]{Passive}
\newacro{r}[R]{Reactive}
\newacro{pr}[PR]{Proactive}
\newacro{tam}[TAM]{Technology Acceptance Model}

\begin{document}
\AddToHookNext{shipout/foreground}{%
\begin{tikzpicture}[remember picture,overlay]
\node[
    anchor=north,
    inner sep=0pt,
    outer sep=0pt
] at ([yshift=6mm]current page.north) {%
    \fbox{%
        \parbox{\dimexpr\textwidth-2\fboxsep-2\fboxrule\relax}{%
            \footnotesize
            \copyright~2026 The Authors. Published by Elsevier Ltd on behalf of IFAC. This is an open access article under the CC BY-NC-ND license. Peer review under the responsibility of the International Federation of Automatic Control.
            }%
    }%
};
\end{tikzpicture}%
}
\begin{frontmatter}

\title{Exploring Human-Robot Collaboration: Analysis of Interaction Modalities in Challenging Tasks}

\author[IDSIA]{Simone Arreghini}
\author[UNIMOREMED]{Cristina Iani}
\author[IDSIA]{Alessandro Giusti}
\author[UNIMOREENG]{Valeria Villani}
\author[UNIMOREENG]{Lorenzo Sabattini}
\author[IDSIA]{Antonio Paolillo}

\address[IDSIA]{Dalle Molle Institute for Artificial Intelligence (IDSIA), USI-SUPSI, Lugano, Switzerland (e-mail: name.surname@idsia.ch)}
\address[UNIMOREMED]{Department of Surgery, Medicine, Dentistry and Morphological Sciences with interest in Transplant, Oncology and Regenerative Medicine, University of Modena and Reggio Emilia, Modena, Italy.}
\address[UNIMOREENG]{Department of Sciences and Methods for Engineering (DISMI), University of Modena and Reggio Emilia, Reggio Emilia, Italy.}

\begin{abstract}
This work compares three interaction modalities for human-robot collaboration: passive, reactive, and proactive. We studied 18 participants assembling a seven-layer colored tower from memory while using nearby and distant blocks. In the passive modality participants worked alone; in the reactive modality a mobile robot helped only upon request; in the proactive modality it initiated brick delivery and error signaling without explicit requests. Although robot assistance increased completion time, most participants preferred collaboration: 67\% preferred proactive behavior and 78\% judged it most useful. These results suggest that timely proactive support can improve user experience in controlled collaborative tasks.
\end{abstract}


\end{frontmatter}

\section{Introduction}\label{sec:intro}
%
Nowadays, robots find applications in diverse social environments such as hotels~\citep{Choi:jhmm:2020}, schools~\citep{Benitti:cae:2012}, and hospitals~\citep{Gonzalez:as:2021}, 
where they assist humans in a variety of collaborative tasks. 
In such scenarios, a robot is not merely asked to execute tasks but needs to collaborate with humans in a socially acceptable way: the robot's presence and behavior must be well perceived by users and operators so that humans and machines can safely and successfully coexist.
This new class of social robots presents novel challenges for researchers in the robotics field. 
Indeed, to engage in meaningful interactions with humans, robots need to understand if nearby humans intend to interact~\citep{Abbate:ras:2023, Arreghini:icra:2024}, interpret human social behavior~\citep{nocentini2019survey}, and possibly predict how a situation may evolve in order to anticipate humans' future actions~\citep{buyukgoz2022two}. 
To build such skills and ensure the efficiency and smoothness of~\ac{hri}, it is crucial to take into account the human side.
In this context, social acceptance can be defined as a positive evaluation that results in the users’ willingness to interact with robots~\citep{dillon2001user}. 
To design robots that meet such requirements, researchers can exploit the mature methods developed in psychology to study collaboration among humans~\citep{Sebanz:tcs:2006} and adapt them to \ac{hri} cases~\citep{clodic2021implement,chan2022argaze}. 

\begin{figure}[!t]
    \centering
    \begin{subfigure}[b]{0.48\columnwidth}
        \includegraphics[trim={12cm 5cm 20cm 15cm},clip,width=\columnwidth]{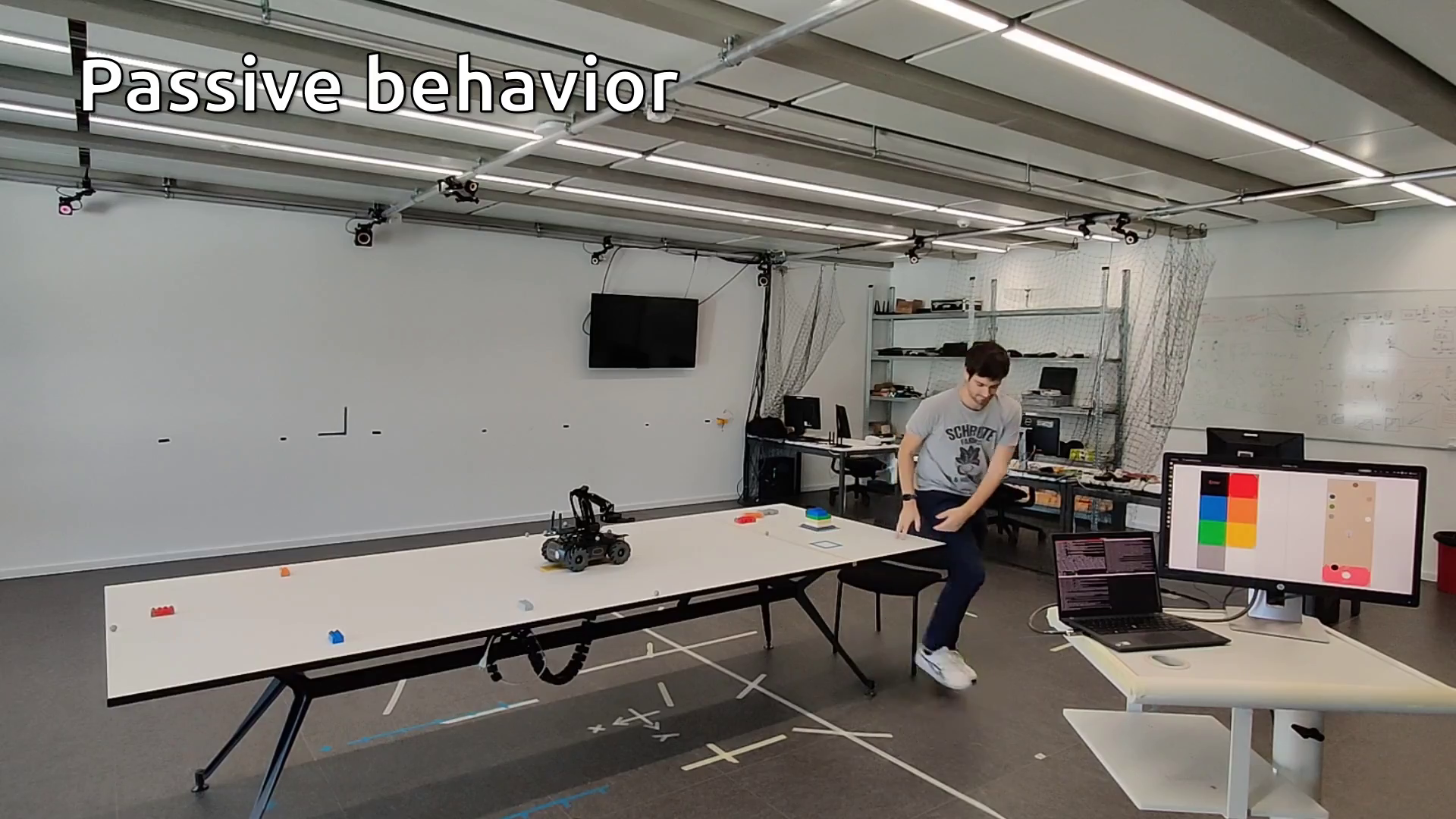}
        \\[5pt]
        \includegraphics[trim={12cm 5cm 20cm 15cm},clip,width=\columnwidth]{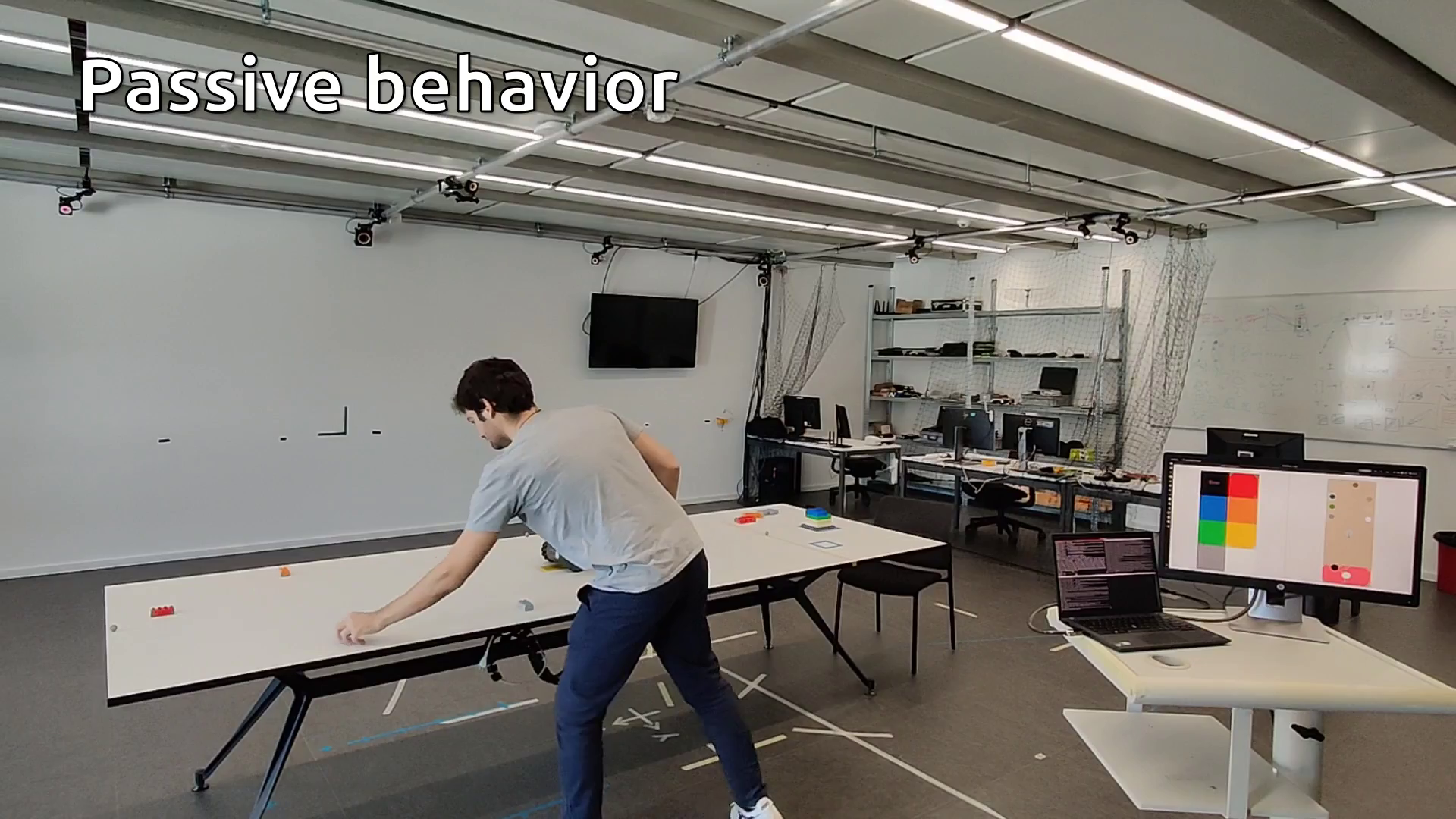}
        \\[5pt]
        \includegraphics[trim={12cm 5cm 20cm 15cm},clip,width=\columnwidth]{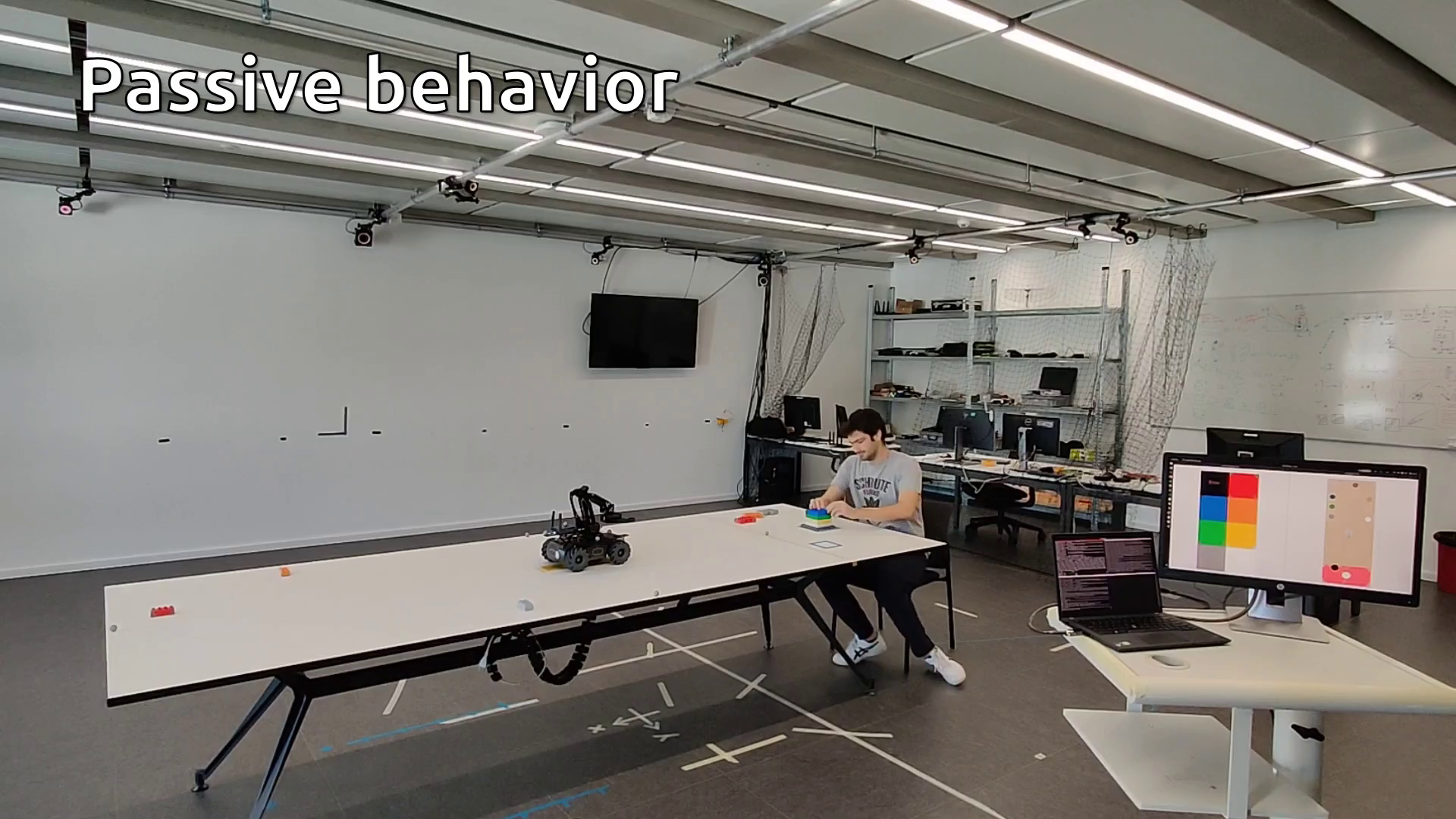}
        \caption{Passive robot}
        \label{fig:passive_rob}
    \end{subfigure}
    ~
    \begin{subfigure}[b]{0.48\columnwidth}
        \includegraphics[trim={10cm 14cm 22cm 6cm},clip,width=\columnwidth]{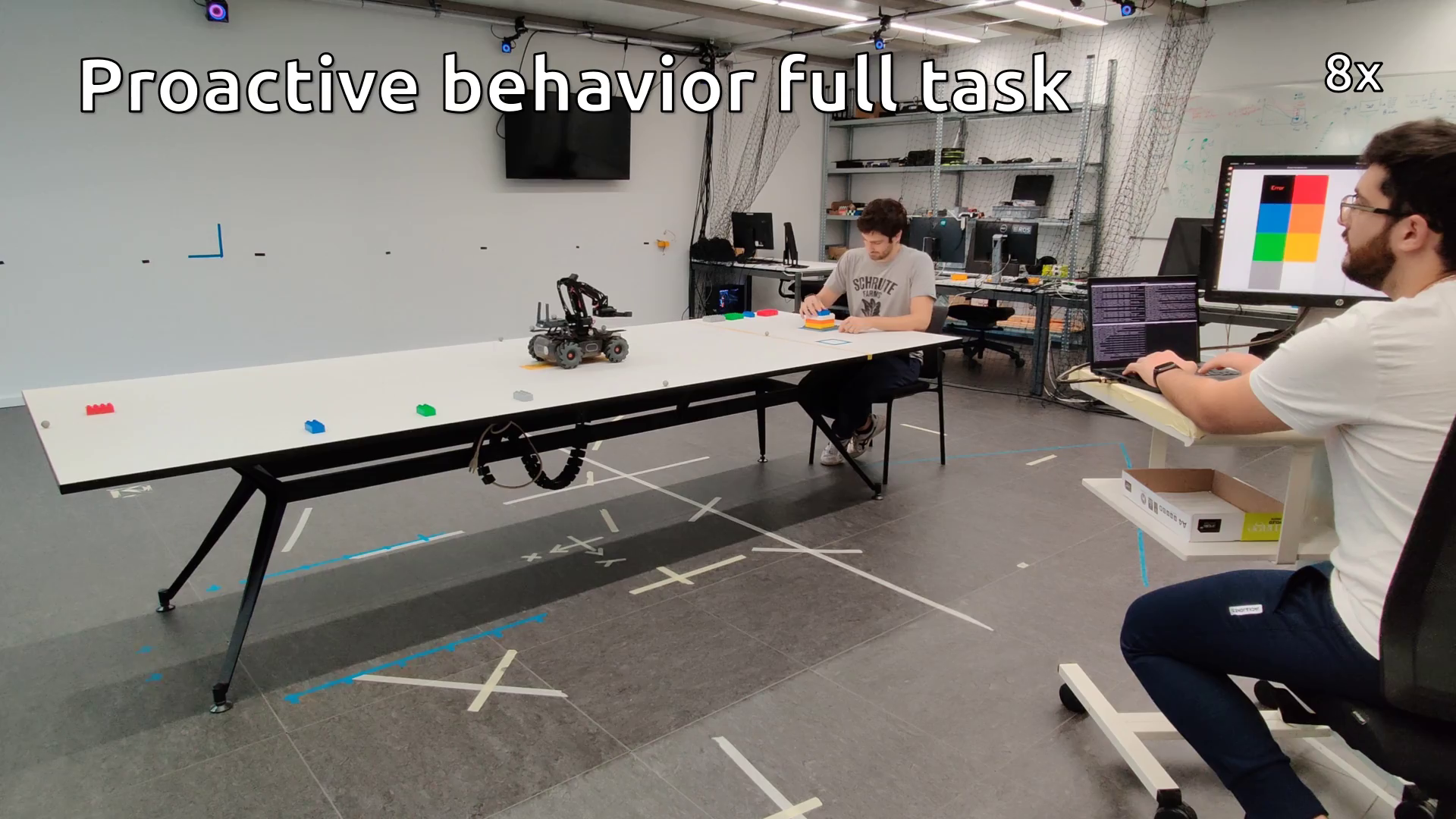}
        \\[5pt]
        \includegraphics[trim={10cm 14cm 22cm 6cm},clip,width=\columnwidth]{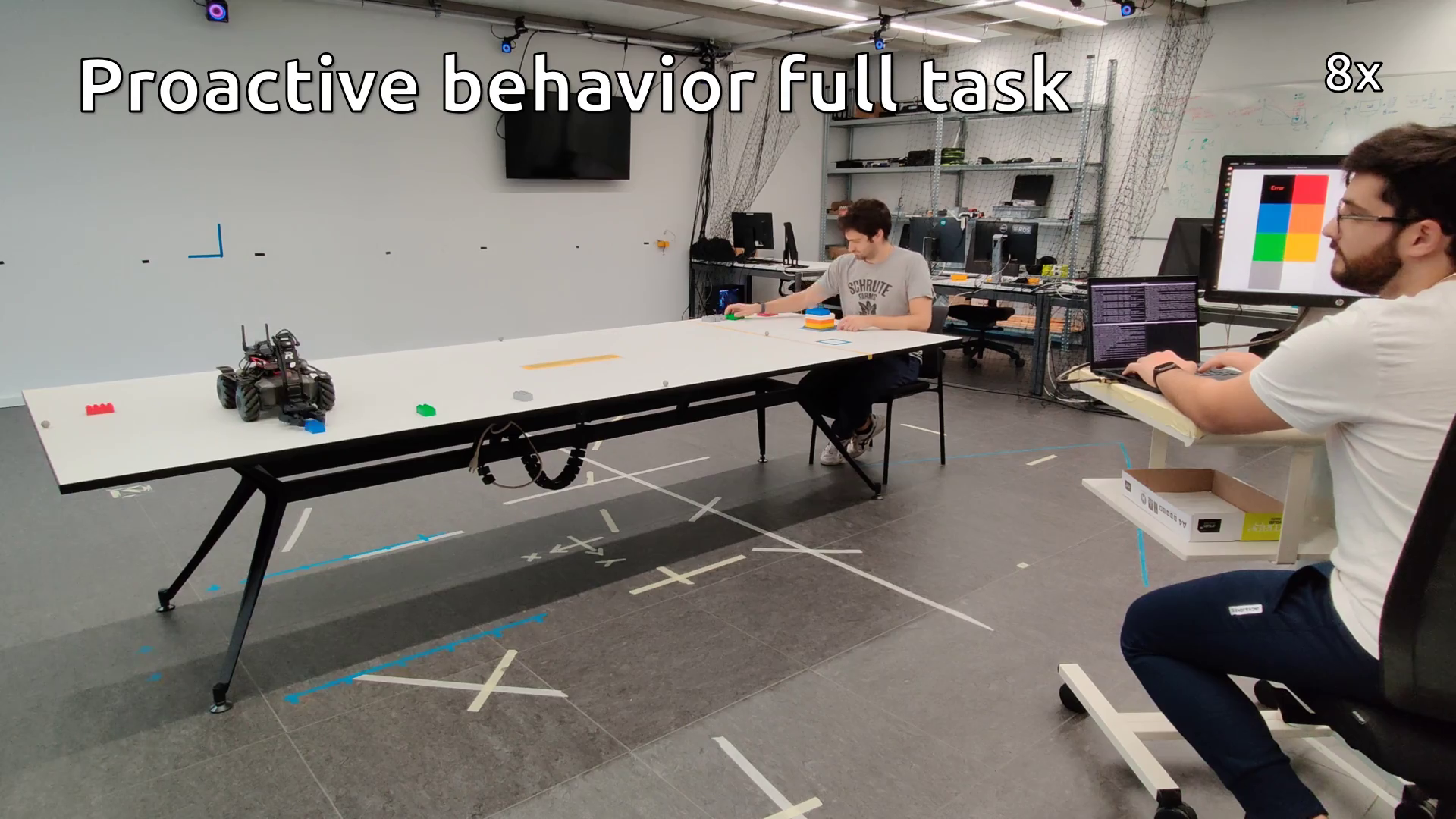}
        \\[5pt]
        \includegraphics[trim={10cm 14cm 22cm 6cm},clip,width=\columnwidth]{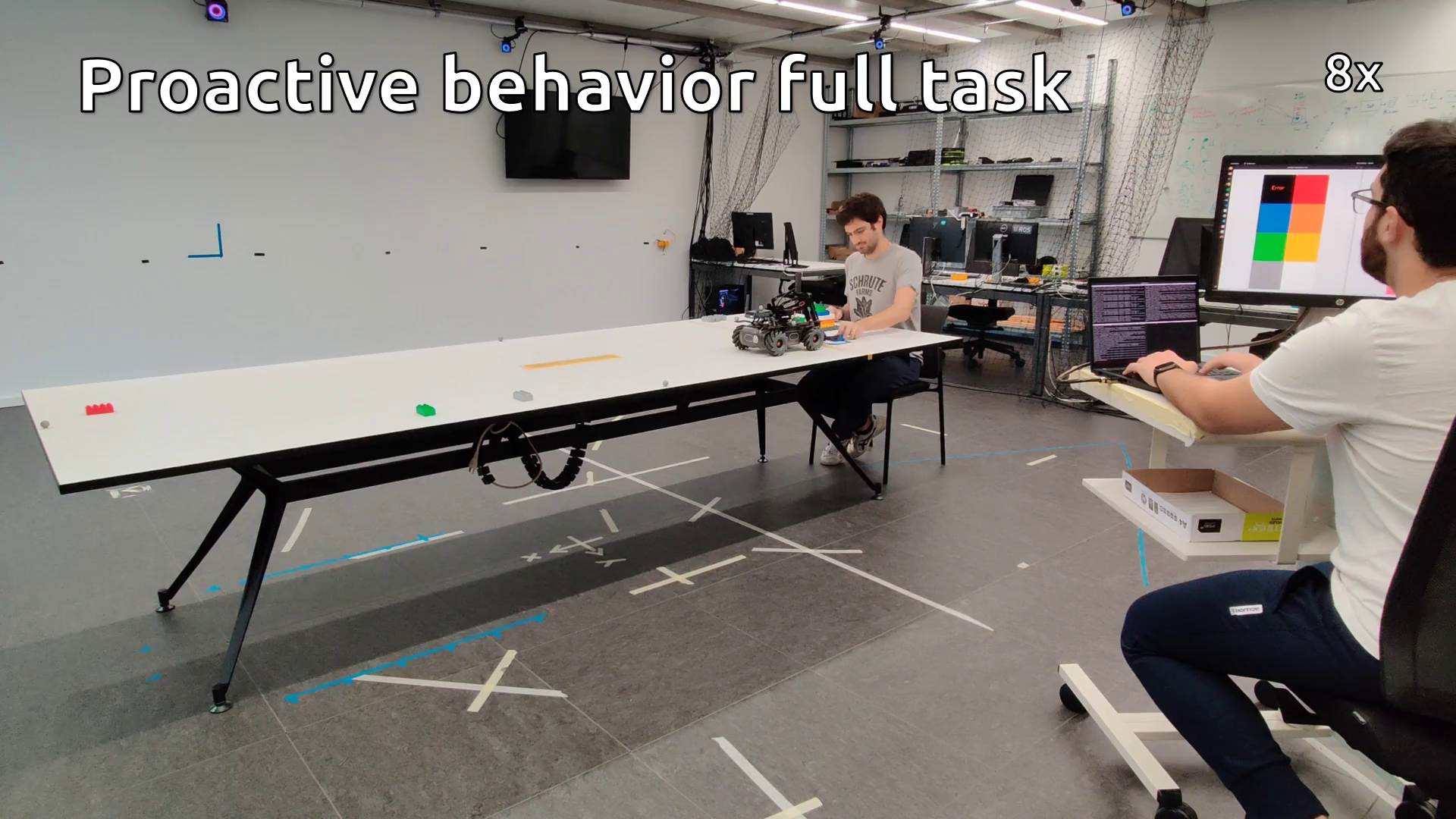}
        \caption{Collaborative robot}
        \label{fig:collab_rob}
    \end{subfigure}
    \caption{In the considered task users built a color-ordered brick tower with (\subref{fig:collab_rob}) or without robot help (\subref{fig:passive_rob}). 
}
    \label{fig:presentation}
\end{figure}

The primary objective of the present study is to examine the psychological processes involved during the execution of collaborative tasks with robots.
Specifically, we investigate both to what extent users embrace the presence of the robot and how they evaluate different interaction modalities, namely reactive and proactive cooperation.
We focus on users' perceptions, preferences, and expectations in the context of human–robot collaboration, as well as on their performance while interacting with the robot.
In particular, we aim to answer the following two questions:
\begin{enumerate}
    \item Which interaction modality do users prefer to establish with a robot? 
    \item What is the impact of a robot's reactive or proactive behavior on users' perception of the task, the robot, and their emotions during the task?
\end{enumerate}
To address these questions, we designed a human–robot collaborative task in which participants built a tower with colored toy bricks (see Fig.~\ref{fig:presentation}). 
They were asked to reproduce a seven-level color sequence shown briefly at the start of each session, making the task cognitively demanding (memorization) and physically challenging (some bricks were placed far from the building area). 
Although simplified, the task abstracts assistive HRI requirements common to healthcare, logistics, and laboratory support: goal memory, object manipulation, reliance on assistance, and user agency.
Each participant completed the task three times: alone (\emph{passive}), with a robot that helped only upon request (\emph{reactive}), and with a robot that proactively intervened (\emph{proactive}). 
%
%
We evaluated: ($i$) task perception, i.e. difficulty, mental and physical demands; ($ii$) robot perception, i.e. usefulness~\citep{davis1989perceived}, reliability, trust~\citep{gaudiello2016trust,rossi2020social,Kraus_2023}; ($iii$) participants’ emotional experience during the interaction.

\section{Related Works}\label{sec:related}
Technological developments have enabled robots to engage in meaningful interactions with users and to be perceived as social entities~\citep{VanDoorn_2017}. 
This development is accelerated by the rapid improvement of Large Language Models~\citep{ye2023improved, frijns2023communication}, while nonverbal communication modalities also play an important role in \ac{hri} processes~\citep{Gasteiger:ijsr:2021,Saunderson:ijsr:2019}.
Robots have to perceive the situation and anticipate human requests based on implicit communication cues, thus displaying emotional intelligence and proactive behavior~\citep{sirithunge2019proactive}. 
 
Multiple studies focused on robot proactivity and aimed at understanding human intention to interact, e.g., analyzing posture~\citep{nocentini2019survey}, gaze~\citep{baraglia2017efficient}, proxemics~\citep{mead2015proxemics} or utterances~\citep{liu2018learning}. 
Increasing the levels of proactivity in social robots can enhance the human perception of anthropomorphic attributes~\citep{tan2020relationship}.
\citet{buyukgoz2022two} proposed a system that integrates intention recognition to enable robots to initiate proactive behavior based on the assessment of human and environmental states. 
\citet{han2019effects} showed that proactive robot behaviors can improve the fluency of physical \acp{hri}. 
Proactivity can benefit~\ac{hri} both on the human side and in terms of robot performance in the context of human-assisted learning~\citep{garrell2013proactive}.
Most of this literature is focused on robot functionalities and only a few studies assessed their impact on human attitudes and perceptions.
Furthermore, even though proactive behavior might appear human-like and beneficial for robot social acceptance, \citet{koelle2020social} also showed that users may perceive it as disturbing or interfering.
The problem of understanding if and when a robot should take initiative in a collaborative task was studied in depth by~\cite{baraglia2017efficient}, within a simple table-top preparation task.
Both behavioral and subjective evaluations of the \ac{hri} quality and the robot’s performance were collected and analyzed.
In detail, the authors compared three different assistive conditions: 
($i$) a human-initiated help condition in which the user decided the timing of the robot’s actions, 
($ii$) a robot-initiated reactive help condition, in which the robot intervened after detecting that the user needed help, and ($iii$) a robot-initiated proactive help condition in which the robot provided help whenever possible. 
They found that people collaborated best with a proactive robot; however, the majority reported preferring to maintain control over when the robot should initiate the intervention. 
The present study builds on~\cite{baraglia2017efficient} by using  a more challenging task (with a larger number of pieces and with the necessity for the user to memorize the task objective), while adding a passive no-assistance baseline. 
Unlike their robot-initiated reactive condition, our \acl{r} modality responds only to explicit requests, whereas our \acl{pr} modality implements the autonomous helping behavior. This lets us assess whether users prefer working alone, retaining initiative, or receiving proactive support.
%
\section{Method}\label{sec:method}
\subsection{Participants}

The study was carried out at the University of Applied Sciences and Arts of Southern Switzerland (SUPSI) with $18$ adults (5 F; $33.3 \pm 8.6$ years), similarly to~\cite{baraglia2017efficient},  and approved by the local ethics committee; informed consent was obtained. 
Seven participants had prior experience with robots but not with ground mobile robots. 
All were randomly assigned to three groups (six each) and exposed to all three modalities—\ac{pa}, \ac{r}, and \ac{pr}, described in detail in Sec.~\ref{sec:methods:exp}—in a Latin square order.
\subsection{Experimental task, design and procedure}\label{sec:methods:exp}
\begin{figure}[!t]
    \centering
    \includegraphics[width=\columnwidth]{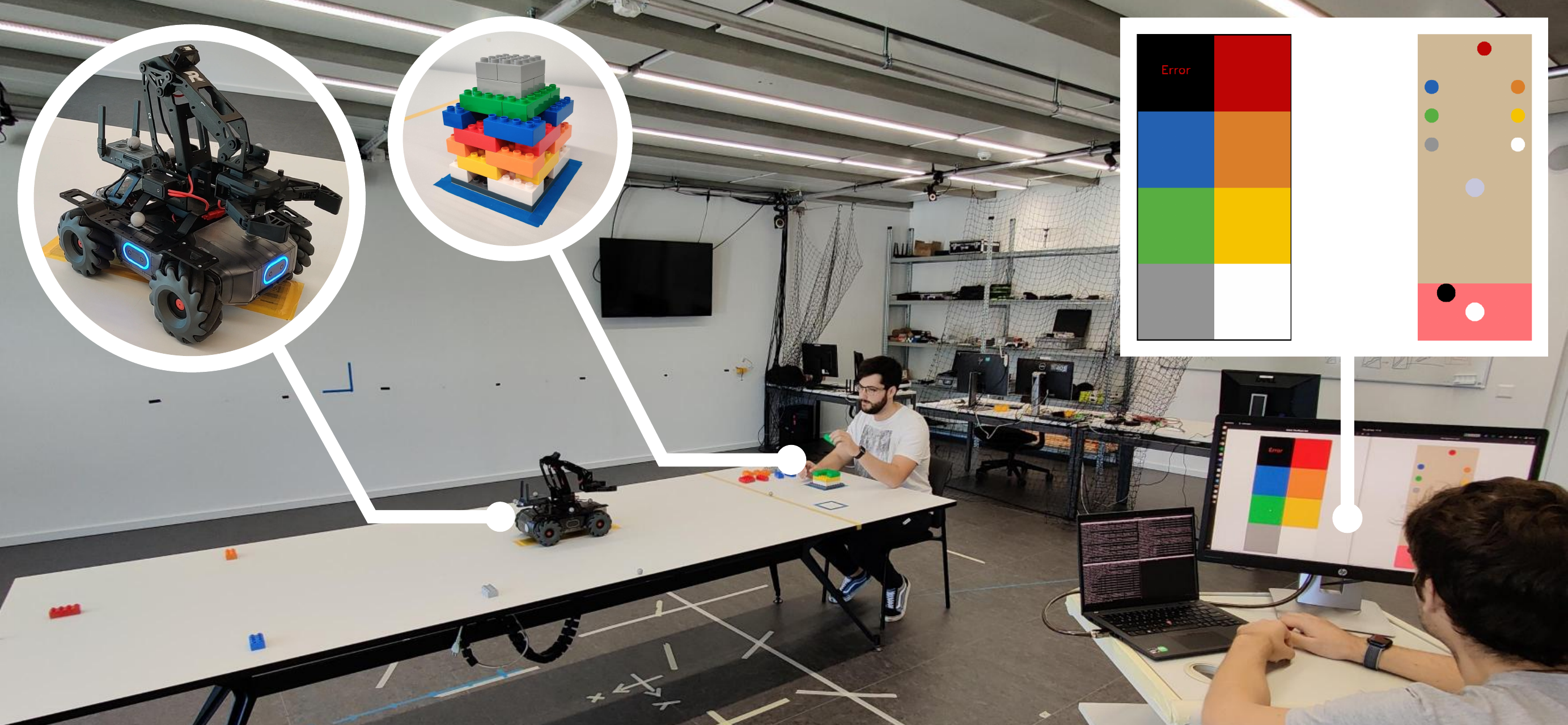}
    \caption{Experimental setup: Users sat at a table and built a color-patterned tower from memory. Some bricks were placed far away, requiring users to stand and retrieve them one at a time. Depending on the modality, a mobile robot assisted by fetching bricks or providing color reminders and error feedback.}
    \label{fig:setup}
\end{figure} 
The experimental setup (Fig.~\ref{fig:setup}) required participants, seated at a table, to build a seven-layer LEGO DUPLO tower, with each layer composed of four bricks of the same color. Only three bricks per layer were within reach; the fourth was placed on the opposite side of the table, requiring participants to stand and retrieve it. 
The task was completed once all bricks were used and any incorrectly colored layer was recorded as an error.
The task was performed under three experimental modalities: passive (\ac{pa}), reactive (\ac{r}), and proactive (\ac{pr}). To maintain cognitive challenge and avoid learning effects, a different color pattern was used in each modality, shown for \SI{30}{s} at the start and then covered. Patterns remained consistent across participants within the same modality.

The \ac{pa} modality served as a baseline to assess task difficulty and users' effort without robotic assistance. 
In \ac{pa}, the robot remained inactive, leaving users to complete the task independently.
In the \ac{r} and \ac{pr} modalities, the robot could assist the user ($i$) by bringing one brick of the requested color from the ones placed on the other side of the table, or ($ii$) by suggesting the correct color to use in the current tower layer. 
The robot conveyed these pieces of information by lighting its LEDs with the corresponding color for a few seconds and pronouncing the color name.
In the \ac{r} modality, the robot provided assistance only upon verbal request: either bringing a brick or suggesting the next color via LED cues and speech. It did not independently detect or notify errors.
In contrast, the \ac{pr} modality featured autonomous robot behavior. After the first brick of a new layer was placed, the robot proactively fetched the fourth brick. The robot also helped users by alerting them to mistakes by flashing red LEDs and saying ``Error''. As in the \ac{r} modality, users could still request color suggestions.
After each modality, participants completed a survey (see Table~\ref{tab:questions}) and estimated their error count. At the end of the experiment, they indicated their preferred and most useful collaboration modality.
The experiment is also presented in the accompanying video\footnote{\url{https://youtu.be/rl3bUhVjC08}}.

\subsection{Robotic platform}
We used a DJI RoboMaster EP Core, a compact mobile robot (\SI{0.32}{m} $\times$ \SI{0.24}{m} $\times$ \SI{0.27}{m}) shown in Fig.~\ref{fig:setup}, equipped with an omnidirectional base, a 2-DOF arm with gripper, RGB camera, LEDs, microphone, and speaker. 
These features enabled both physical assistance and verbal/visual communication during the task. 
The robot’s linear speed and angular velocity were limited for safety and comfort reasons to \SI{2}{m/s} and \SI{5}{rad/s}, respectively. 
Its position was tracked using an external motion capture system. 
The platform’s small size and friendly appearance facilitated safe and non-threatening interaction, even for users unfamiliar with robots. The system was controlled via ROS2 and a dedicated low-level module.
\subsection{Wizard-of-Oz implementation}
To simplify robot perception and control, we used a Wizard-of-Oz setup~\citep{wizard_of_oz}, with an operator discreetly teleoperating the robot from a desk \SI{3}{m} away as shown in Fig.~\ref{fig:setup}. 
Participants were told that the operator was present for monitoring, and were unaware of the teleoperation, perceiving the robot as fully autonomous.
The operator used two custom GUIs: one for triggering brick pick-and-place actions via a table-like interface showing brick positions and key locations (idle, pickup, drop-off), and another for issuing visual and audio feedback (e.g., color cues, error alerts) via corresponding LED and speech commands. 
This setup enabled smooth behavior aligned with the experimental modalities and also isolated user responses to assistance timing from perception and planning robustness.
\subsection{Evaluation metrics}\label{sec:methods:evaluation}

To understand the impact of the robot’s interaction modality, in addition to recording performance metrics, we conducted a fine-grained assessment of how participants perceived the task, the robot, and the interaction with it. 
We also evaluated the feelings that users experienced during task execution~\citep{eagly1998attitude,Naneva_2020}.
Indeed, acceptance can be considered as a continuous decision process affected by the experiences gathered while performing a task~\citep{koelle2020social}, which in turn affects what users think (cognitive component) and feel (emotional component), as pointed out by~\cite{Naneva_2020} and \cite{nomura2006experimental}.

\subsubsection{Performance metrics}
To compare participants' performance in completing the task under the three experimental modalities, we collected several objective metrics during task execution: 
task completion time (in seconds), number of user errors, explicit brick requests, explicit color suggestion requests, and robot-initiated error feedback events.
\subsubsection{Subjective metrics}

Inspired by previous studies ~\citep{baraglia2017efficient, akalin2022you, hoffman2019evaluating}, we designed a questionnaire with 25 statements rated on a 7-point Likert scale (1: strongly disagree, 7: strongly agree).
The statements, reported in Tab.~\ref{tab:questions}, assessed: 
($i$) task perception in terms of difficulty, physical and mental demand, time pressure (statements $1$-$4$), 
($ii$) emotional experience during the task (statements $5$-$9$), 
and ($iii$) perception of the robot  in terms of appearance, behavior, trust, usefulness and interaction quality (statements $10$-$25$). 
After the \ac{pa} modality, participants answered only items 1–9 plus a question about whether they would have preferred to receive help. 
After the \ac{r} and \ac{pr} modalities, they completed the full questionnaire. Thus, the $10$th statement has two different versions between the \ac{pa} and the other two modalities (10a and 10b in Tab.~\ref{tab:questions}, respectively).
\begin{table*}[!tb]
    \centering
    \caption{Average and standard deviation of the scores collected in our user study.}
    \begin{tabular}{clcccc} 
        \toprule
        \multicolumn{2}{c}{\sc{Statements}} & PA & R & PR & \sc{Type of Score}\\
        \midrule
        1 & The task was easy to accomplish & $6.1 \pm 1.4$ & $6.3 \pm 0.8$ & $6.8 \pm 0.5$ & Task\\
        2 & The task was mentally demanding & $3.1 \pm 2.0$ & $2.8 \pm 1.8$ & $2.4 \pm 1.7$ & Task\\
        3 & The task was physically demanding & $3.2 \pm 2.0$ & $1.7 \pm 1.4$ & $1.6 \pm 1.2$ & Task\\
        4 & I felt pressure in accomplishing the task & $3.1 \pm 2.0$ & $2.9 \pm 2.0$ & $2.7 \pm 2.0$ & Task\\
        5 & I felt in control of the outcome & $5.8 \pm 1.7$ & $6.2 \pm 1.1$ & $6.0 \pm 1.7$ & Emotional\\
        6 & During the task, I felt anxious & $2.5 \pm 1.8$ & $2.5 \pm 1.8$ & $1.9 \pm 1.1$ & Emotional\\
        7 & During the task, I felt comfortable & $5.9 \pm 1.5$ & $6.1 \pm 1.4$ & $6.2 \pm 1.2$ & Emotional\\
        8 & During the task, I felt bored & $2.4 \pm 1.6$ & $2.0 \pm 1.5$ & $2.3 \pm 1.6$ & Emotional\\
        9 & I was accurate in performing the task & $6.1 \pm 1.4$ & $6.4 \pm 0.9$ & $6.7 \pm 0.6$ & Emotional\\
        10a & I would have preferred to receive some help & $5.3 \pm 2.0$ & - & - & - \\
        10b & The robot was helpful in accomplishing the task & - & $6.4 \pm 0.7$ & $6.3 \pm 1.3$ & Robot\\
        11 & The robot and I equally contributed to the task & - & $4.5 \pm 2.1$ & $4.8 \pm 2.2$ & Robot\\
        12 & The robot's contribution to the task was higher & - & $3.2 \pm 2.2$ & $3.1 \pm 2.2$ & Robot\\
        13 & The interaction with the robot felt natural & - & $5.8 \pm 1.5$ & $6.4 \pm 0.9$ & Robot\\
        14 & The robot promptly reacted to my requests & - & $6.1 \pm 1.4$ & $6.2 \pm 1.1$ & Robot\\
        15 & The robot was accurate in perceiving my needs/actions/commands & - & $6.1 \pm 1.4$ & {$6.7 \pm 0.5$}&{Robot}\\
        16 & The robot’s behavior met the needs of the task & - & $5.9 \pm 1.6$ & $6.4 \pm 1.3$ & Robot\\
        17 & The robot was too slow & - & $2.6 \pm 1.7$ & $2.9 \pm 2.0$ & Robot\\
        18 & I felt in competition with the robot & - & $1.2 \pm 0.6$ & $1.3 \pm 0.5$ & Robot\\
        19 & I felt the robot was reliable & - & $5.7 \pm 1.6$ & $6.2 \pm 1.0$ & Robot\\
        20 & I trusted the robot & - & $6.2 \pm 1.2$ & $6.5 \pm 0.6$ & Robot\\
        21 & The robot’s behavior was easy to understand & - & $6.4 \pm 1.2$ & $6.7 \pm 0.6$ & Robot\\
        22 & I liked the robot’s appearance & - & $6.1 \pm 1.1$ & $6.0 \pm 1.2$ & Robot\\
        23 & I felt comfortable in interacting with the robot & - & $6.1 \pm 1.3$ & $6.6 \pm 0.5$ & Robot\\
        24 & I felt safe in interacting with the robot & - & $6.4 \pm 0.9$ & $6.7 \pm 0.5$ & Robot\\
        25 & I am willing to interact again with the robot & - & $6.2 \pm 1.4$ & $6.2 \pm 1.4$ & Robot\\
        \bottomrule
    \end{tabular}
    \label{tab:questions}
\end{table*}
\subsubsection{Holistic scores}
Given the distinct aims of the questionnaire statements, we grouped them into three domains: task perception (statements $1$–$4$), emotional experience ($5$–$9$), and perception of the robot ($10$–$25$). 
We calculated composite scores for each domain by averaging participant responses, applying inversion to reversed statements ($2$, $3$, $4$, $6$, $8$, $17$, and $18$) as needed. 
These resulting metrics are referred to as the \emph{task}, \emph{emotional}, and \emph{robot} scores.
%
\section{Results}\label{sec:results}

Statistical analyses were performed with IBM SPSS v.28. 

\subsection{Performance evaluation}
\begin{figure}[!tb]
    \centering
    \includegraphics[width=\columnwidth]{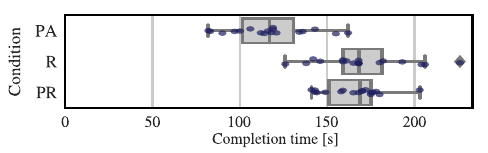}
    \caption{Completion time was significantly lower in the \ac{pa} modality, with no difference between \ac{r} and \ac{pr}.}
    \label{fig:time_plots}
\end{figure}

We analyzed task completion time across the three modalities using a repeated-measures ANOVA, with modality as a within-participant factor and execution order as a between-participant factor ($\alpha = 0.05$). As shown in Fig.~\ref{fig:time_plots}, there was a significant main effect of modality, $F(2,30) = 38.60$, $p < 0.001$: participants performed the task fastest in the \ac{pa} modality, with no significant difference between \ac{r} and \ac{pr} ($p = 0.47$). Neither the order of execution nor its interaction with modality was significant ($F < 1$ and $F(4,30) = 1.37$, $p = 0.27$). 
Longer assisted times likely reflect interaction overhead: users waited for robot motions and feedback, whereas in \ac{pa} they retrieved bricks continuously.
Other performance measures revealed a marked contrast between interaction styles. In the \ac{r} modality, $88\%$ of participants requested the robot’s help for brick delivery at least five times, and $50\%$ asked for color suggestions at least once. Conversely, in \ac{pr}, only one participant explicitly asked for help, as most relied on the robot’s autonomous feedback, which was triggered in $22\%$ of sessions (up to three times per session).
Participants’ self-assessment of errors was inaccurate: they overestimated errors in \ac{pa} modality ($12$ predicted vs. $6$ actual) and underestimated them in  \ac{r} modality ($6$ vs. $10$).
\subsection{Subjective evaluation}
Table~\ref{tab:questions} shows mean responses and standard deviations across modalities. Using Wilcoxon signed-rank tests, we compared ratings across \ac{pa}, \ac{r}, and \ac{pr} modalities. Only statements $1$–$9$ were analyzed for \ac{pa} due to the absence of robot interaction.
Participants rated the task as significantly easier in \ac{pr} than \ac{pa} (statement $1$, $Z = -2.57$, $p = 0.01$), and perceived higher physical effort in \ac{pa} than in \ac{pr} ($Z = -3.09$, $p = 0.002$), with a similar trend for \ac{pa} compared with \ac{r} ($Z = -2.73$, $p = 0.06$), for statement $3$. This suggests higher perceived physical effort when participants performed the task without the help of the robot.
The \ac{pr} modality also trended toward higher accuracy (statement $9$, $Z = -1.81$, $p = 0.07$) and lower anxiety (statement $6$, $Z = -1.69$, $p = 0.09$) than \ac{pa}.
Between \ac{r} and \ac{pr}, the \ac{pr} modality scored higher for task ease (statement $1$, $Z = -2.12$, $p = 0.034$), and precision of robot perception (statement $15$, $Z = -2.14$, $p = 0.04$). Weaker trends favoring \ac{pr} were also seen for natural interaction (statement $13$, $p = 0.054$), task fit (statement $16$, $p = 0.06$), and comfort (statement $23$, $p = 0.07$).
Overall, participants found the robot helpful and did not perceive it as competitive, regardless of interaction modality.
\subsection{Holistic evaluation}
\begin{figure}[!t]
    \centering
    \includegraphics[width=\columnwidth]{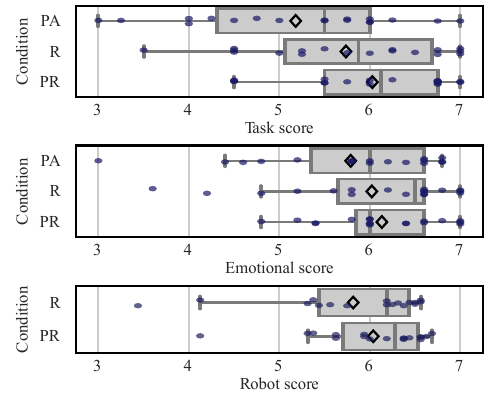}
    \caption{Distribution of the task (top), emotional (center), and robot score (bottom) for the three modalities; dots and diamonds are individual samples and means.}
    \label{fig:scores_over_modality}
\end{figure}
\begin{figure}
    \centering\includegraphics{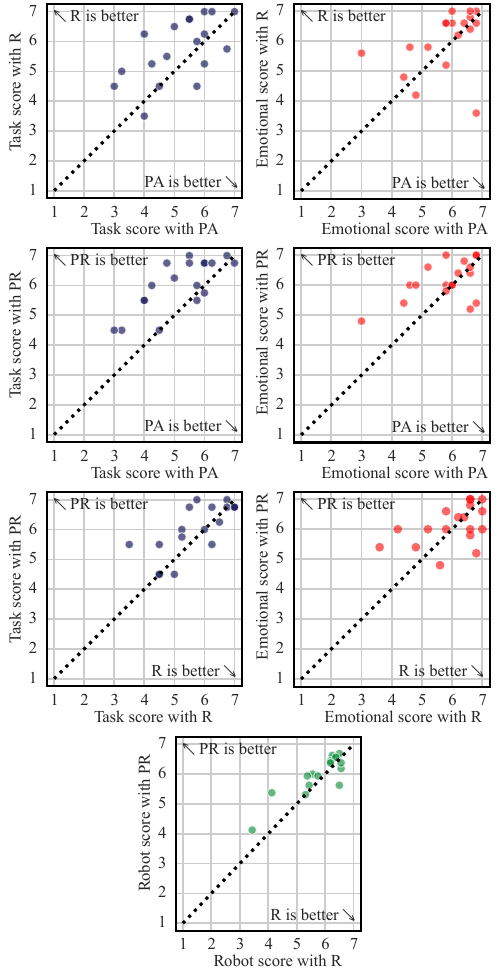}
    \caption{Scores scatter plots between modalities. 
    }
    \label{fig:scores_comparison}
\end{figure}
The holistic task, emotional, and robot scores (Sec.~\ref{sec:methods:evaluation}) were analyzed using Wilcoxon signed-rank tests; means and standard deviations are reported in Fig.~\ref{fig:scores_over_modality}. Task scores in \ac{pa} were significantly lower than in both \ac{r} ($Z = -2.13$, $p = 0.03$) and \ac{pr} ($Z = -3.21$, $p = 0.001$), reflecting reduced perceived difficulty with robotic support. Notably, $61\%$ of participants expressed a desire for assistance following the \ac{pa} session.
Emotional scores exhibited a marginal improvement in \ac{pr} compared to \ac{pa} ($Z = -1.68$, $p = 0.09$), suggesting enhanced positive affect during proactive interaction. Additionally, robot scores were significantly higher in \ac{pr} than in \ac{r} ($Z = -1.98$, $p = 0.05$), indicating a preference for proactive robotic behavior.
Figures~\ref{fig:scores_over_modality} and~\ref{fig:scores_comparison} depict the trends across holistic scores. Task and emotional scores were more favorable in robot-assisted modalities, particularly in \ac{pr}.
Overall, $67\%$ of participants preferred \ac{pr}, and $78\%$ considered it the most useful. Among the remaining participants, $28\%$ favored \ac{r}, with only one individual selecting \ac{pa}. Those preferring \ac{r} cited a desire for greater control, consistent with findings of~\cite{baraglia2017efficient}.
%
\section{Discussion and Conclusion}\label{sec:conclusions}

This study evaluated how different robot interaction modalities (passive, reactive, and proactive) affect human performance, perceptions, and emotional responses during a cognitively and physically demanding collaborative task. Compared to performing the task alone (passive), both reactive and proactive robot behaviors reduced perceived task difficulty and physical effort, with the proactive modality showing the most positive impact.

Our results partially align with~\cite{baraglia2017efficient}: whereas users there preferred controlling robot intervention timing, most participants in our study favored proactive support.
We attribute this difference to our setup (e.g. greater cognitive load of memorizing color sequences, physical challenge of brick retrieval, and more natural interaction flow made possible by the Wizard-of-Oz) which spared participants from manual task logging.

Interestingly, despite longer task completion times, 
participants reported a more positive task experience in both assisted modalities. This suggests that users are willing to trade efficiency for reduced effort and enhanced interaction quality. The robot was perceived as helpful and non-intrusive in both reactive and proactive modalities, with no significant feelings of loss of control or competition. 
Nevertheless, individual preferences varied: while most users favored proactive assistance, a subset preferred reactive behavior due to a desire for greater control. This motivates adaptive interaction modalities for diverse user expectations and collaboration styles.
The results should be interpreted in light of the small sample size, albeit consistent with prior work~\citep{baraglia2017efficient}. Future studies should use larger, more diverse samples to improve generalizability and support analyses of individual differences (e.g. prior robot experience).

Overall, our results support the design of assistive robots that proactively engage with users while maintaining transparency and user agency. Future work should investigate whether these insights hold in more complex scenarios, such as  dynamic environments or tasks requiring autonomous navigation, and across different robotic platforms. Such research will aim to develop socially intelligent, context-aware human-robot collaboration.

\begin{ack}
This work was supported by the European Union through (project SERMAS) and the Swiss State Secretariat for Education, Research and Innovation (contract 22.00247).
\end{ack}

\bibliography{biblio}

@string{acc = "IEEE Access"}

@string{aes = "AI \& Society"}

@string{ar = "Adv. Rob."}

@string{as = "Appl. Sci."}

@string{auro = "Auton. Robot."}

@string{ce = "Comp. \& Education"}

@string{chb = "Comp. in Human Behavior"}

@string{chi = "ACM Conf. on Human Factors in Comp. Sys."}

@string{ehfe = "Enc. of human factors and ergonomics"}

@string{frai = "Frontiers in Rob. and AI"}

@string{icra = "IEEE Int. Conf. Robot. and Autom."}

@string{ijhcs = "Int. J. Human-Computer Studies"}

@string{ijrr = "Int. J. Robot. Res."}

@string{ijsr = "Int. J. of Social Robot."}

@string{iros = "IEEE/RSJ Int. Conf. Intelligent Robots Sys."}

@string{hri = "ACM/IEEE Int. Conf. Human-Robot Int."}

@string{hsp = "The Handbook of Soc. Psych."}

@string{jhmm = "J. of Hospitality Marketing \& Management"}

@string{jhri = "J. of Human-Robot Int."}

@string{jsr = "J. of Service Research"}

@string{misq = "MIS Quarterly"}

@string{ras = "Robot. Auton. Syst."}

@string{robotics = "Robotics"}

@string{roman = "IEEE Int. Symp. Robot and Human Int. Comm."}

@string{thms = "IEEE Trans. on Human-Machine Syst."}

@string{thri = "ACM Trans. on Human-Robot Int."}

@string{tcs = "Trends in Cogn. Sci."}

@inproceedings{Arreghini:icra:2024,
 author = {S. Arreghini and G. Abbate and A. Giusti and A. Paolillo},
 title = {Predicting the Intention to Interact with a Service Robot: the Role of Gaze Cues},
 booktitle=icra, 
 pages = {993--999},
 year = {2024},
}

@article{Saunderson:ijsr:2019,
  title={How robots influence humans: A survey of nonverbal communication in social human--robot interaction},
  author={Saunderson, Shane and Nejat, Goldie},
  journal=ijsr,
  volume={11},
  pages={575--608},
  year={2019}
}

@article{Gasteiger:ijsr:2021,
  title={Factors for personalization and localization to optimize human--robot interaction: A literature review},
  author={Gasteiger, Norina and Hellou, Mehdi and Ahn, Ho Seok},
  journal=ijsr,
  pages={1--13},
  year={2021}
}

@article{hoffman2019evaluating,
  title={Evaluating fluency in human-robot collaboration},
  author={Hoffman, Guy},
  journal=thms,
  volume={49},
  number={3},
  pages={209--218},
  year={2019}
}

@article{akalin2022you,
  title={Do you feel safe with your robot? Factors influencing perceived safety in human-robot interaction based on subjective and objective measures},
  author={Akalin, Neziha and Kristoffersson, Annica and Loutfi, Amy},
  journal=ijhcs,
  volume={158},
  pages={102744},
  year={2022}
}

@article{nocentini2019survey,
  title={A survey of behavioral models for social robots},
  author={Nocentini, Olivia and Fiorini, Laura and Acerbi, Giorgia and Sorrentino, Alessandra and Mancioppi, Gianmaria and Cavallo, Filippo},
  journal=robotics,
  volume={8},
  number={3},
  pages={54},
  year={2019},
  publisher={MDPI}
}

@article{eagly1998attitude,
  title={Attitude structure},
  author={Eagly, Alice and Chaiken, Shelly},
  journal=hsp,
  volume={1},
  pages={269--322},
  year={1998}
}

@article{baraglia2017efficient,
  title={Efficient human-robot collaboration: when should a robot take initiative?},
  author={Baraglia, Jimmy and Cakmak, Maya and Nagai, Yukie and Rao, Rajesh PN and Asada, Minoru},
  journal=ijrr,
  volume={36},
  number={5-7},
  pages={563--579},
  year={2017}
}

@article{Sebanz:tcs:2006,
  title={Joint action: bodies and minds moving together},
  author={Sebanz, Natalie and Bekkering, Harold and Knoblich, G{\"u}nther},
  journal=tcs,
  volume={10},
  number={2},
  pages={70--76},
  year={2006}
}

@article{Choi:jhmm:2020,
author = {Youngjoon Choi, Miju Choi, Munhyang (Moon) Oh and Seongseop (Sam) Kim},
title = {Service robots in hotels: understanding the service quality perceptions of human-robot interaction},
journal = jhmm,
volume = {29},
number = {6},
pages = {613-635},
year = {2020}
}

@article{Gonzalez:as:2021,
  title={Social robots in hospitals: a systematic review},
  author={Gonz{\'a}lez-Gonz{\'a}lez, Carina Soledad and Violant-Holz, Ver{\'o}nica and Gil-Iranzo, Rosa Maria},
  journal=as,
  volume={11},
  number={13},
  pages={5976},
  year={2021}
}

@article{Abbate:ras:2023,
 author = {G. Abbate and A. Giusti and V. Schmuck and O. Celiktutan and A. Paolillo},
 title = {Self-Supervised Prediction of the Intention to Interact with a Service Robot},
 journal = ras,
 volume = {171},
 pages = {104568},
 year = {2024}
}

@article{Benitti:cae:2012,
  title={Exploring the educational potential of robotics in schools: A systematic review},
  author={Benitti, Fabiane Barreto Vavassori},
  journal=ce,
  volume={58},
  number={3},
  pages={978--988},
  year={2012}
}

@article{Kraus_2023,
  title={On the Role of Beliefs and Trust for the Intention to Use Service Robots: An Integrated Trustworthiness Beliefs Model for Robot Acceptance},
  author={Kraus, Johannes and Miller, Linda and Klumpp, Mariel{\`e}ne and Babel, Franziska and Scholz, David and Merger, Julia and Baumann, Martin},
  journal=ijsr,
  pages={1--24},
  year={2023}
}

@article{Naneva_2020,
  title={A systematic review of attitudes, anxiety, acceptance, and trust towards social robots},
  author={Naneva, Stanislava and Sarda Gou, Marina and Webb, Thomas L and Prescott, Tony J},
  journal=ijsr,
  volume={12},
  number={6},
  pages={1179--1201},
  year={2020}
}

@article{VanDoorn_2017,
  title={Domo arigato Mr. Roboto: Emergence of automated social presence in organizational frontlines and customers’ service experiences},
  author={Van Doorn, Jenny and Mende, Martin and Noble, Stephanie M and Hulland, John and Ostrom, Amy L and Grewal, Dhruv and Petersen, J Andrew},
  journal=jsr,
  volume={20},
  number={1},
  pages={43--58},
  year={2017}
}

@article{wizard_of_oz,
  title={Wizard of {Oz} studies in {HRI}: a systematic review and new reporting guidelines},
  author={Riek, Laurel D},
  journal=jhri,
  volume={1},
  number={1},
  pages={119--136},
  year={2012}
}

@article{davis1989perceived,
  title={Perceived usefulness, perceived ease of use, and user acceptance of information technology},
  author={Davis, Fred D},
  journal=misq,
  pages={319--340},
  year={1989}
}

@inproceedings{rossi2020social,
  title={How social robots influence people’s trust in critical situations},
  author={Rossi, Alessandra and Dautenhahn, Kerstin and Koay, Kheng Lee and Walters, Michael L},
  booktitle=roman,
  pages={1020--1025},
  year={2020}
}

@article{ye2023improved,
  title={Improved trust in human-robot collaboration with {ChatGPT}},
  author={Ye, Yang and You, Hengxu and Du, Jing},
  journal=acc,
  year={2023},
  volume={11},
  pages={55748-55754}
}

@article{sirithunge2019proactive,
  title={Proactive robots with the perception of nonverbal human behavior: A review},
  author={Sirithunge, Chapa and Jayasekara, AG Buddhika P and Chandima, DP},
  journal=acc,
  volume={7},
  year={2019}
}

@article{liu2018learning,
  title={Learning proactive behavior for interactive social robots},
  author={Liu, Phoebe and Glas, Dylan F and Kanda, Takayuki and Ishiguro, Hiroshi},
  journal=auro,
  volume={42},
  pages={1067--1085},
  year={2018}
}

@inproceedings{mead2015proxemics,
  title={Proxemics and performance: Subjective human evaluations of autonomous sociable robot distance and social signal understanding},
  author={Mead, Ross and Matari{\'c}, Maja J},
  booktitle=iros,
  year={2015}
}

@article{buyukgoz2022two,
  title={Two ways to make your robot proactive: Reasoning about human intentions or reasoning about possible futures},
  author={Buyukgoz, Sera and Grosinger, Jasmin and Chetouani, Mohamed and Saffiotti, Alessandro},
  journal=frai,
  volume={9},
  pages={929267},
  year={2022}
}

@article{dillon2001user,
  title={User acceptance of information technology},
  author={Dillon, Andrew},
  journal=ehfe,
  volume={1},
  pages={1105--1109},
  year={2001}
}

@misc{koelle2020social,
  title={Social Acceptability in HCI: A Survey of Methods, Measures, and Design Strategies},
  author={Koelle, Marion and Ananthanarayan, Swamy and Boll, Susanne},
  journal=chi,
  year={2020}
}

@article{clodic2021implement,
  title={What Is It to Implement a Human-Robot Joint Action?},
  author={Clodic, Aurelie and Alami, Rachid},
  journal={Robotics, AI, and humanity: Science, ethics, and policy},
  pages={229--238},
  year={2021}
}

@article{chan2022argaze,
    title={Design and Implementation of a Human-Robot Joint Action Framework using Augmented Reality and Eye Gaze},
    author={Chan, Wesley P and Crouch, Morgan and Hoang, Khoa and Chen, Charlie and Robinson, Nicole and Croft, Elizabeth},
    journal = thri,
    year = {2022}
}

@article{frijns2023communication,
  title={Communication models in human--robot interaction: an asymmetric MODel of ALterity in human--robot interaction},
  author={Frijns, Helena Anna and Sch{\"u}rer, Oliver and Koeszegi, Sabine Theresia},
  journal=ijsr,
  volume={15},
  number={3},
  pages={473--500},
  year={2023}
}

@article{nomura2006experimental,
  title={Experimental investigation into influence of negative attitudes toward robots on human--robot interaction},
  author={Nomura, Tatsuya and Kanda, Takayuki and Suzuki, Tomohiro},
  journal=aes,
  volume={20},
  pages={138--150},
  year={2006}
}

@article{gaudiello2016trust,
  title={Trust as indicator of robot functional and social acceptance. An experimental study on user conformation to {iCub} answers},
  author={Gaudiello, Ilaria and Zibetti, Elisabetta and Lefort, S{\'e}bastien and Chetouani, Mohamed and Ivaldi, Serena},
  journal=chb,
  volume={61},
  pages={633--655},
  year={2016}
}

@article{tan2020relationship,
  title={Relationship between social robot proactive behavior and the human perception of anthropomorphic attributes},
  author={Tan, Hao and Zhao, Ying and Li, Shiyan and Wang, Wei and Zhu, Ming and Hong, Jie and Yuan, Xiang},
  journal=ar,
  volume={34},
  number={20},
  pages={1324--1336},
  year={2020}
}

@inproceedings{han2019effects,
  title={The effects of proactive release behaviors during human-robot handovers},
  author={Han, Zhao and Yanco, Holly},
  booktitle=hri,
  pages={440--448},
  year={2019}
}

@inproceedings{garrell2013proactive,
  title={Proactive behavior of an autonomous mobile robot for human-assisted learning},
  author={Garrell, Anais and Villamizar, Michael and Moreno-Noguer, Francesc and Sanfeliu, Alberto},
  booktitle=roman,
  pages={107--113},
  year={2013}
}

\end{document}